\pdfoutput=1
\documentclass[11pt]{article}

\usepackage{ACL2023}
\usepackage{times}
\usepackage{latexsym}

\usepackage{nert}

\renewcommand{\finalversion}[1]{\unskip}

\usepackage[T1]{fontenc}
\usepackage[utf8]{inputenc}

\usepackage{microtype}

\usepackage{inconsolata}

\usepackage{makecell}

\title{Lost in Translationese? Reducing Translation Effect \\Using Abstract Meaning Representation}

\author{Shira Wein \\
Georgetown University \\
  \texttt{sw1158@georgetown.edu} \\\And
  Nathan Schneider \\
Georgetown University \\
  \texttt{nathan.schneider@georgetown.edu} \\}

\begin{document}
\maketitle
\begin{abstract}
Translated texts bear several hallmarks distinct from texts originating in the language.
Though individual translated texts are often fluent and preserve meaning, at a large scale, translated texts have statistical tendencies which distinguish them from text originally written in the language (``translationese'') 
% The presence of translationese in training or test sets can affect model performance, but mitigating the effect of translationese in human translated text is understudied.
and can affect model performance.
We frame the novel task of \emph{translationese reduction} and 
hypothesize that Abstract Meaning Representation (AMR), a graph-based semantic representation which abstracts away from the surface form, can be used as an interlingua to reduce the amount of translationese in translated texts. By parsing English translations into an AMR and then generating text from that AMR, the result more closely resembles originally English text across three quantitative macro-level measures, without severely compromising fluency or adequacy. 
%We verify that the resulting text, in addition to have reduced translationese, does not have severely comprimised fluency or adequacy by applying four NLG automatic metrics and performing a thorough human evaluation study.
We compare our AMR-based approach against three other techniques based on machine translation or paraphrase generation.
%, which make use of machine translation, T5-based paraphrasing, and BART-based paraphrasing. 
This work makes strides towards reducing translationese in text and highlights the utility of AMR as an interlingua. 
\end{abstract}

\section{Introduction}

The term \emph{translationese} \citep{gellerstam1986translationese} describes the features unique to translated texts:
% such as when a translation feels less fluent than a text originally written in that language \citep{baker2019corpus,de2017towards}
% \nss{but emphasize: other things as well}. 
% “Identifiable features that may be related to the nature of the translation activity itself” (Olohan, 2004)
% Language affected by translationese exhibits
the specific syntactic and semantic patterns found in human translations \citep{Teich2003,volanskyetal2015}.
When the presence of translationese is not addressed in training or test sets, evaluation scores can be overinflated \citep{zhang-toral-2019-effect,graham-etal-2020-statistical,wang-etal-2023-understanding}, model performance can be impacted \citep{yu2022translate,ni-etal-2022-original}, or system-generated output can be dispreferred by humans \citep{freitag-etal-2019-ape}.
However, if used correctly, actively leveraging translated texts in language model training can lead to improved performance in machine translation systems
% \nss{so translationese is sometimes good if used correctly?} 
% \citep{parthasarathi-etal-2021-sometimes-want,kurokawa-etal-2009-automatic,lembersky-etal-2011-language,lembersky-etal-2012-adapting,twitto-etal-2015-statistical}. 
\citep{parthasarathi-etal-2021-sometimes-want,kurokawa-etal-2009-automatic,lembersky-etal-2012-adapting,twitto-etal-2015-statistical}.

\begin{figure}[t]
\begin{small}
\textbf{Original translation:} Now, however, he is to go before the courts once more because the public prosecutor is appealing.
\smallbreak
% \small
\textbf{Parsed AMR:}
\begin{verbatim}
(c / contrast-01
    :ARG2 (g / go-02
        :ARG0 (h / he)
        :ARG4 (c2 / court)
        :mod (a / again
            :mod (o / once))
        :time (n / now)
        :ARG1-of (c3 / cause-01
            :ARG0 (a2 / appeal-02
                :ARG0 (p / person
                   :ARG0-of (p2 / prosecute-01)
                   :ARG1-of (p3 / public-02))))))
\end{verbatim}
\smallbreak
\end{small}
\small\textbf{Generated sentence:} But now he will go to court once again because the public prosecutor is appealing.
% \end{small}
  \caption{Example of our ``parse-then-generate'' approach to mitigating translationese, which involves first translating the sentence %(from \citet{nisioi-etal-2016-corpus}) 
  into an AMR and then generating back into a sentence.}
  % \nss{possible to add arrows to show it's a pipeline?}
  \label{fig:process}
\end{figure}

% \sw{need to add more here about why this task matters and how it's novel}
Previous work has studied the characteristics and impact of translationese.\footnote{Though the term ``translationese'' is still commonly used in NLP\slash MT, it is less commonly used in translation studies \citep{jimenez-crespo-2023-translationese}. In this work, we use the term to refer to specific characteristics which may arise out of the translation process, not necessarily corresponding to unnaturalness in the text \citep{kunilovskaya-lapshinova-koltunski-2019-translationese}.} In this work, we set out to reduce the amount of translationese in human-translated text while preserving the meaning.
%which is a notably understudied task.
This corresponds to a task of automatic \emph{translationese reduction} for human translations (\cref{sec:new_task}).
This task is important given the effect of translationese in both training and test sets, and is relevant to automatic tools for post-editing translations.

We hypothesize that translationese can be reduced using a formal semantic representation as an interlingua, because the representation abstracts away from the surface form while ensuring the integrity and continuity of the core elements of meaning.
Specifically, we explore the utility of the Abstract Meaning Representation \citep[AMR;][]{banarescu-etal-2013-abstract} formalism as an interlingua/intermediate representation for this task. We introduce a ``parse-then-generate'' technique
% uses AMR as an intermediate representation, 
which takes a text affected by translationese, parses that text into an AMR, and then generates text which is more like original English from that AMR. 
% This straightforward approach uses AMR as an interlingua to abstract away from surface-level features, such as translation effect.
% \nss{previous paragraph explains AMR is an interlingua}

In addition to our proposed ``parse-then-generate'' technique leveraging AMR, we experiment with two additional promising techniques. First, given the similarity between our task of translationese reduction and paraphrase generation, we apply two paraphrase models (one T5-based and one BART-based) to translationese reduction. We suspect that these models should also reduce the effect of translation on the surface form and lead to reduced explicitation, which is a hallmark of translationese \citep{baker1993corpus,gellerstam1996translations}. Next, given the promise of ``back-translation'' for this task and the distinct set of translationese features appearing in machine versus human translations \citep{bizzoni-etal-2020-human}, we test whether back-translation using machine translation actually reduces the amount of human translationese (\cref{sec:methods}). 

% Additionally, we experiment with two alternative approaches to translationese reduction, one based on round-trip machine translation, and one based on syntactically controlled generation. 

We assess the performance of each technique for translationese reduction through experimentation with three macro-level translationese metrics (\cref{sec:metrics}), an automatic evaluation of meaning preservation using three NLG metrics (\cref{ssec:automatic_adequacy}), a thorough human evaluation of both fluency and adequacy, and qualitative analysis of the output (\cref{ssec:human_evaluation}).
% Our three automatic metrics for translationese reduction are: (1)~type-token ratio (TTR), (2)~presence of cohesive markers, and (3)~unigram bag-of-part-of-speech.

While AMR generation does not produce perfectly fluent texts (as judged by human evaluators), we find that the AMR-based approach is the only method which aids in translationese reduction across all metrics while preserving sufficient adequacy and fluency, highlighting the promise of AMR as an interlingua. The code for the AMR parse-then-generate technique and our evaluation protocol is available at \url{https://github.com/shirawein/amr-translationese}.

\section{Background on Translationese}
\label{sec:background}
% \nss{I feel like there is a missing piece before experimental setup: motivation and overview of the method (in more detail than the intro). E.g., give an example or two of translationese from a known source language (figure 1?). Mention the sorts of properties that translationese is known to have. Discuss why it would be good to reduce the effects of translationese}

Translated and non-translated text (originally written in that language) exhibit various differences referred to as ``translationese'' \citep{gellerstam1986translationese}. 
Translated text is often less lexically rich, has simpler constructions, exhibits explicitation, and demonstrates specific lexical and word order choices \citep{baker1993corpus,gellerstam1996translations}.
An example exhibiting translationese can be seen at the top of \cref{fig:process}.
% Translationese
% -afflicted text is often dispreferred by human readers \citep{freitag-etal-2019-ape}, and can impact downstream tasks trained on it \citep{yu2022translate,ni-etal-2022-original}.
% \footnote{Related work on translationese is discussed in \cref{sec:related_work}.}
% Machine translations and human translations exhibit different types of translationese \citep{bizzoni-etal-2020-human}.
% \nss{how does this relate to ?}
The presence of translationese is not necessarily indicative of a low-quality translation \citep{kunilovskaya-lapshinova-koltunski-2019-translationese}, 
and prior work has shown that human raters are not able to accurately predict whether text is translated or not \citep{tirkkonen2002translationese,wein-2023-human}.

Two basic types of translationese include:
(1)~interference from the source, such as the presence of syntactic patterns typical of the source language \citep{Teich2003}, and (2)~over-normalizing to the target language, for example not translating abnormalities seen in the source text.
The patterns and characteristics of translationese vary by mode and register, most notably if the translation is written or spoken \citep{bernardini2016epic}; translationese found in human translations versus machine translations (MT) also exhibit different characteristics \citep{bizzoni-etal-2020-human}.

% \nss{after they alter sentence embeddings, do they generate a supposedly reduced-translationese sentence, and use it in a downstream task? or do they only care about the embeddings?}\nss{for the evaluations that they do have, are any of them appropriate as baselines for the AMR approach?}
% \sw{which was the first study to remove the effect of translationese in embeddings}

Related work has also considered the impact and causes of translationese via investigating the algorithmic biases which lead to translationese in MT \citep{vanmassenhove-etal-2021-machine}, avoiding the influence of translationese in training and testing by means of translationese classifiers and zero-shot multilingual MT %to produce translations which seem original in both the source and target languages 
\citep{riley-etal-2020-translationese}, and exploring the utility of word-by-word glosses in producing fluent translations 
% developing an MT system which first generated rough glosses of the original text and then translated the resulting gloss into a fluent \sw{native-like translation} \
\citep{pourdamghani-etal-2019-translating}.

Prior work has developed automatic classifiers of translationese, which detect whether the text exhibits translationese or not \citep{rabinovich-wintner-2015-unsupervised,rabinovich-etal-2017-personalized,pylypenko-etal-2021-comparing}.
A couple of studies have sought to counteract the effects of translationese.
Contemporaneously to the present work, \citet{jalota-etal-2023-translating} evaluated translationese classifier accuracy before and after applying style transfer to translated texts.
%Work contemporaneous with ours, also related to the markers of translationese in text, set out to assess the quality of translation-based style transfer via translationese classifier accuracy \citep{jalota-etal-2023-translating}.
%Related recent work also removed translationese implicitly encoded in vector embeddings \citep{dutta-chowdhury-etal-2022-towards}.
In a similar vein, \citet{dutta-chowdhury-etal-2022-towards} removed translationese implicitly encoded in vector embeddings (but did not produce a reduced-translationese version of the translated text).
% Ours is the first work explicitly aiming to reduce the amount of translationese in human-translated text.
Our work is novel in that we (1)~frame the task of translationese reduction as one which reduces the statistical patterns of translationese, while preserving meaning and fluency, (2)~introduce three methods of translationese reduction, and (3)~demonstrate on both quantitative and qualitative metrics that our AMR-based approach succeeds at reducing the presence of translationese.

\section{Translationese Reduction}
\label{sec:new_task}
% Because of the decrease in naturalness in translationese-affected text, reducing the amount of translationese in human translations (or after machine translation) would allow translated texts to sound more native-like, and could be used after translation prior to incorporation in a test set.
We undertake this task of automatic translationese reduction for English, where the input is a sentence that has been translated into English and the output is a paraphrase that better resembles a sentence that originated in English.
We do not assume access to the source sentence that was translated, or even to the source language.

We formulate the task of translationese reduction by proposing automatic metrics for diminishing the hallmarks of translationese, informed by prior work documenting the notable features of translated texts.

Importantly, fluency and adequacy must be preserved in the task of translationese reduction, as conveying the same meaning is paramount. Thus, the reduction of translationese hallmarks across various automatic metrics may not come at the cost of adequacy or fluency, and any viable method for translationese reduction needs to maintain these aspects of the text while reducing features of translationese.

In this work, we approach translationese reduction by first mapping the translated English into a meaning representation in order to abstract away from superficial aspects of expression that may be artifacts of the translation process. 
This meaning representation is intended as an intermediary, or ``interlingua,'' between the two ``dialects'' of English: translationese and originally English text.
% non-translationese.
 For example, in \cref{fig:process}, we see that we start with a translation, parse the text into an AMR graph, and generate from that AMR graph a sentence more like original English text.

\section{Methods}
\label{sec:methods}
First, in \cref{ssec:data}, we introduce the data that we use for our experiments. Next, in the three subsections that follow, we outline the three approaches we develop to take on our task of translationese reduction: one using paraphrase generation models (\Cref{ssec:paraphrase}); one using machine translation (\Cref{ssec:mt}); and the third approach using AMR as an interlingua (\Cref{ssec:amr_motivation}).\footnote{We also developed and experimented with an approach using syntactically controlled generation, adapting the model from \citet{chen-etal-2019-controllable}. However we found that this produced nonsensical output, as was the case for even the example generated sentences in \citeposs{chen-etal-2019-controllable} paper. Thus, we have omitted this method from our results.}

\subsection{Data}
\label{ssec:data}
For our experiments, we use the \emph{English corpus of Native, Non-native and Translated Texts} (ENNTT) \citep{nisioi-etal-2016-corpus}, which is based on Europarl data \citep{koehn-2005-europarl}. ENNTT consists of three distinct (non-parallel) sets of data: translated text, text originally in English uttered by non-native speakers, and text originally in English uttered by native speakers.
% \nss{is the source language/L1 relevant?} \sw{not really because it's indicated in the fact that it's europarl}
The translated texts are edited versions of the transcriptions, not real-time translations. To create the English Europarl proceedings\slash translated dataset, the spoken utterances were (1)~transcribed, then (2)~edited by the original speaker, then (3)~translated by a human native speaker of English \citep{nisioi-etal-2016-corpus}.
Here, we use 2000 sentences from the translated and native datasets: 1000 translated sentences and 1000 originally English sentences uttered by native speakers.
% From these we produce new generated sentences, first\nss{?} parsed into AMRs. 
We use the originally English datasets
% \footnote{Note that by ``natively English'' in this context, we are referring to the native language of the \emph{source text}, as the translations into English also involve native speakers of English.}
to compare the part-of-speech values of translated English versus original English in \cref{ssec:unigram_pos}.
% We use both the translated and native sentences so that we can collect baseline statistics of the originally English sentences, aiming to determine whether our technique makes the translated sentences less affected by translationese.

%First, we test out using machine translation as an interlingua and find that this does not reduce the amount of translationese using the outlined metrics. Next, we leverage syntactically controlled generation to produce generated text with the same meaning but controlled syntax. This approach also does not result in translationese reduction.
%After briefly describing the first two approaches, we describe the third method, which uses AMR as an interlingua and successfully mitigates the presence of translationese.
% \nss{don't say here which methods worked. the point is to present the ideas, and results will come later}

\subsection{Paraphrase Generation}
\label{ssec:paraphrase}

%Given the relevance of paraphrase generation to our novel task of translationese reduction, it is intuitive to use existing state-of-the-art paraphrase models as initial approaches to our task. This is because, we are ultimately in search of a natural paraphrase to a text afflicted by translationese.
Given that our goal of translationese reduction is a form of paraphrasing---producing a meaning-preserving alternative phrasing that better resembles originally English text---we experiment with two preexisting paraphrase models. 
% Note that these were trained to generate generic paraphrases;
We examine whether the produced paraphrases reduce the effects of translationese.
% In our experimentation, we use two models for paraphrase generation, one which 

\paragraph{Para T5.} The first is a T5-based paraphrase model \citep{chatgpt_paraphraser}\footnote{\url{https://huggingface.co/humarin/chatgpt\_paraphraser\_on\_T5\_base}},  
trained on the ChatGPT paraphrase dataset \citep{chatgpt_paraphrases_dataset}. The model is based on the T5-base model and uses transfer learning to combine the benefits of the ChatGPT paraphrases and the paraphrases generated from this model. There are 420,000 items in the training data, with each consisting of a question and five paraphrases produced by ChatGPT.\footnote{We use the AutoTokenizer pretrained from the \texttt{chatgpt\_paraphraser\_on\_T5\_base} model as well as the pretrained \texttt{chatgpt\_paraphraser\_on\_T5\_base} AutoModelForSeq2SeqLM.}

\paragraph{Para BART.} The second paraphrase system 
% \nss{paraphrase citation separate from the BART citation?} \sw{none available}
uses BART \citep{lewis2019bart} \footnote{\url{https://huggingface.co/eugenesiow/bart-paraphrase}}.
%The BART-based paraphrase model 
It was trained by fine-tuning a pretrained seq2seq (text2text) \texttt{bart-large} model on the Quora \footnote{\url{https://www.kaggle.com/c/quora-question-pairs}}, PAWS \citep{zhang-etal-2019-paws}, and MSR paraphrase corpora \citep{dolan2005automatically}.
The Quora Question Pair dataset consists of 404,290 rows; the PAWS (Paraphrase Adversaries from Word Scrambling) corpus consists of 751,450 rows; the MSR (Microsoft Research) paraphrase corpus consists of 5,800 pairs of sentences.
All three datasets consist of paraphrase candidate pairs (the MSR dataset has a human annotation indicating whether the sentences are paraphrases).\footnote{We use the \texttt{BartForConditionalGeneration} pretrained model and the BARTTokenizer.}

\subsection{Round-Trip Machine Translation}
\label{ssec:mt}

The next approach uses round-trip machine translation through a second language. 
This approach is motivated by prior work which explored back- and forward-translation as a tool for identifying data which is original to the target language (not the source language).
\Citet{riley-etal-2020-translationese} found that including back-translated data in translation models leads to a minor improvement in BLEU score.
Round-trip machine translation has also been found to aid grammatical error correction under some conditions \citep{kementchedjhieva-sogaard-2023-grammatical}; this further motivates the use of machine translation in human translationese reduction, given that the features of translationese in human and machine translation are distinct \citep{bizzoni-etal-2020-human}).
% \nss{rephrased}
% \sw{add \citep{kementchedjhieva-sogaard-2023-grammatical}: \url{https://aclanthology.org/2023.findings-eacl.165/}}
Because of prior work leveraging back- and forward-translation related to improving naturalness and identifying translationese, we suspect that this approach might aid in the reduction of characteristics of human translationese.
% \nss{The intuition is that perhaps the MT system will be subject to different biases than a human translator, so that using a second language as proxy for the meaning, and then returning to the original language, might smooth out some of the human translationese characteristics.}

Using the EasyNMT package\footnote{\url{https://github.com/UKPLab/EasyNMT}}, we take the original English text which is afflicted by translationese, translate it into French, and then translate the French back into English. We use French because it is a Europarl language and EN-FR machine translation is of high quality.

\subsection{Abstract Meaning Representation}
\label{ssec:amr_motivation}

Our third and primary approach is to use semantic parsing to abstract away from the phrasing of the translation while maintaining meaning. 
We use the Abstract Meaning Representation formalism as the intermediate semantic representation because it captures the core elements of meaning while de-centering the specific phrasing associated with sentences. AMR encapsulates the meaning of a sentence in a rooted, directed graph.
Each node in the graph corresponds to a semantic unit in the sentence, and is labeled with an entity or event type (``concept''). Edges between nodes reflect relationships between semantic units.
We hypothesize that AMR is an especially good choice to serve as an interlingua in the reduction of translationese because it abstracts away from the surface form to isolate the semantic elements of the sentence.
As function words, inflectional morphology, specific word order and word choice are not captured in the AMR, this could help deal with issues such as unnatural phrasing and promote lexical richness.
% Even among a variety of semantic representations (such as UCCA \cite{abend-rappoport-2013-universal}
% or PropBank \citep{palmer_2005_propbank}),
% \nss{PropBank is used in AMR though}
% \nss{incomplete}
Further, the abundance of work on text-to-AMR parsing and AMR-to-text generation means that the quality of output is relatively high compared to other semantic representations.

Upon the translated and (distinct) originally English sentences, we apply our ``parse-then-generate'' method: (1)~parse the sentence into an AMR, then (2)~from the parsed AMR, generate a sentence. This process is illustrated in \cref{fig:process}. We make use of the \texttt{amrlib}\footnote{\url{https://github.com/bjascob/amrlib}} BART-based text-to-graph AMR parser and T5-based graph-to-text generator.
% \nss{also cite the publication describing the model} \sw{there is none}

To determine the effectiveness of using AMR as an interlingua to abstract away from translation effect, we apply three translationese metrics to see if the parsed-then-generated sentences have characteristics more similar to the originally English sentences than the translated sentences.
% \finalversion{We apply four translationese metrics to see if the translated sentences then have characteristics more similar to the native sentences after using AMR as an interlingua.}
% We apply four translationese metrics to see if the translated p-then-g sentences have characteristics more similar to the native than the translated sentences.

\section{Measuring Translationese}
\label{sec:metrics}

Prior work has established several statistical properties of translated text \citep{volanskyetal2015}.
Measures known to distinguish translations from non-translations include:
%To measure the ability of systems to reduce the presence of translationese, we use three quantitative measures, drawing from prior work, enumerated below: 
(1)~type-token ratio (TTR), (2)~the presence of cohesive markers, and (3)~unigram bag-of-part-of-speech (POS) tags.
Note that while the metrics we apply here are informed by prior work both in natural language processing and translation studies, these metrics show a partial picture of the range of statistical patterns observed in translated texts.
% \sw{add more discussion of the partial nature of the (established) translationese metrics provided more space in the camera-ready.}
These are not ``translation universals,'' per se, so much as they are statistical tendencies \citep{jimenez-crespo-2023-translationese} observed in prior work on features of translated texts.

We compare system outputs on these metrics, using the original translations as a baseline, to assess whether each system successfully mitigates the observed presence of translationese.
% \nss{I don't quite follow what is added here:}
% Note that, in addition to the motivation for the use of each metric, \citet{volanskyetal2015} find that TTR, markers of cohesion, interference from the source language via part-of-speech, and the presence of function words are accurate predictors of translationese.
In each subsection, we detail the metric as well as the results for each approach with that metric.

\subsection{Type-token Ratio}
\label{ssec:ttr}
\textbf{Type-token ratio (TTR)}, as used by \citet{rabinovich-etal-2016-similarities}, quantifies lexical richness. Lower TTR reflects \emph{text simplification}, in which a sentence in the source language has fewer linguistically complex features upon translation into the target language \citep{blum1978universals}. 
\Citet{vanmassenhove-etal-2021-machine} find a decrease in lexical richness in text affected by translationese.

% \nss{discussion should follow order of rows in the table}
Type-token ratio results can be found in \cref{tab:metric_quant}.
First, for our AMR parse-then-generate approach, the type-token ratio results point to success in reducing translationese. Type-token ratio, and thus lexical complexity, \emph{increases} as expected once we apply our AMR parse-then-generate approach to the translated sentences. 
The AMR-based technique increases type-token ratio to 0.1002.
%which is not as drastic as the change caused by the BART-based paraphrase model but still an improvement.
The BART-based paraphrase model also successfully reduces the presence of translationese and leads to an even more drastic change, improving the type-token ratio to 0.1172.

However, we find that when applying the machine translation back-translation technique, type-token ratio decreases from 0.0890 to 0.0850, indicating further diminished linguistic complexity.
Similarly, the T5-based paraphrase model diminishes lexical complexity and the type-token ratio is reduced to 0.0736.

\begin{table}[t]
\centering
\small
\begin{tabular}{ r c@{~}l >{\hspace{4em}}c@{~}l}
 \toprule
& \multicolumn{2}{c}{\textbf{TTR} ($\uparrow$)} & \multicolumn{2}{c}{\textbf{Cohesive Markers} ($\downarrow$)} \\ [0.5ex] 
 \midrule
Translations & 0.0890 & & 461 \\
\midrule
% Natives & 0.1075 & 325 \\
% Natives P-then-G & 0.1170 & 280 \\
MT & 0.0850 &  & 483 \\
Para BART & 0.1172 & \chk & 277 & \chk \\
Para T5 & 0.0736 & & 446 & \chk \\
AMR P-then-G & 0.1002 & \chk & 348 & \chk \\
 \bottomrule
\end{tabular}
\caption{Type-token Ratio (TTR) and number of cohesive markers for the 1000 translated sentences before and after using each of the translationese reduction methods. ``MT'' indicates MT back-translation and ``AMR P-then-G'' is an abbreviation for AMR Parse-then-Generate. \chk~indicates improvement over the baseline of the original translation.
% \nss{is bolding necessary?}\nss{would it make sense to add Originally English like Table 2?}
}
\label{tab:metric_quant}
\end{table}

% \begin{figure*}[htb]
% \centering
% \includegraphics[trim={0cm 0cm 0cm 0cm},clip,width=0.75\textwidth]{scatter.pdf}
% \caption{Scatterplot with trendlines the dataset before and after the application of each technique, showing Universal Dependencies average summed distance for every sentence length---for every sentence length, we average all of the UD summed distances.
% \sw{need to edit this to remove the Natives data and add in MT and Paraphrase Results}
% }
% \label{fig:udasd}
% \end{figure*}

\subsection{Cohesive Markers}
\label{ssec:cohesive_markers}
\textbf{Cohesive markers} are sentence transitions like ``besides,'' ``in other words,'' and ``furthermore.''
They are often overused in translations \citep{rabinovich-etal-2016-similarities}, consistent with the explicitation hypothesis \citep{blum1986shifts}, which suggests that information implied or understood in an originally English text is often specified in translations. The presence of cohesive markers in the ENNTT corpus, which we use in this work, is investigated in \citet{rabinovich-etal-2016-similarities}.
% \nss{what is a cohesive marker? give some examples}
We would expect the presence of cohesive markers to decrease when successfully reducing the amount of translationese in a text.

In the case of cohesive markers, the MT back-translation technique again \emph{exacerbates} translationese, with the number of cohesive markers increasing from 461 to 483. 
Both paraphrase models, on the other hand, reduce the number of cohesive markers: the T5-based paraphrase model produces a small decrease (from 461 to 446), while the BART-based paraphrase model leads to a much more drastic change (from 461 to 277).
% Using the AMR parse-then-generate techniques, the number of cohesive markers
% %, which are characteristic of translated text, 
% also decreases as expected once we apply our approach 

The AMR parse-then-generate approach also successfully reduces the number of cohesive markers
(from 461 to 348).
Some cohesive markers are captured in the parsed AMRs (such as \texttt{contrast} being used to mark ``however'' in \cref{fig:process}), while cohesive markers which carry less meaning are not captured. This results in only information-carrying cohesive markers being included in the generated text, whereas less critical cohesive markers (which may be products of translationese) are omitted.
% \nss{what are the implications of some being captured and others not?} 
% Again, AMR parse-then-generate leads to translationese reduction, though not as drastic a change as that caused by the BART-based paraphrase model.
% % being represented. 

\subsection{Unigram Bag-of-POS}
\label{ssec:unigram_pos}
\textbf{Unigram bag-of-POS} measures source interference on grammatical structure \citep{pylypenko-etal-2021-comparing,volanskyetal2015}. As supported by the shining through hypothesis \citep{Teich2003}, the grammatical structure (as approximated by part-of-speech (POS) \emph{n}-grams) of translationese-affected text should be more similar to that of the source language than text originally written in the target language.
% \sw{add another sentence here to clarify}
In order to collect part-of-speech tags for our test data, use the spaCy \texttt{en\_core\_web\_sm} part-of-speech tagger.\footnote{\url{https://spacy.io/models/en}}

When using the AMR parse-then-generate approach, the unigram bag-of-POS results 
% are broken down into quantity and proportion and are included in 
suggest that our approach decreases the proportion of key tags.
% \nss{what do they tell us?}
\Citet{pylypenko-etal-2021-comparing} show that the POS tag relative frequency of ADV (adverbs) can be a predictor of the presence of translationese, perhaps as well as the relative frequency of determiners and adpositions.
For all tags, the highest \emph{number} of tags for most part-of-speech tags (12 out of 17) appears in the translated text. This is because the sentences are longer for the translated sentences than any other data,
% \footnote{The number of tokens in 1000 translated sentences is 32,596 in total; after parsing-then-generating the total number of tokens in the 1000 originally English sentences is 25,499.}, 
likely due to explicitation.
The number of tokens in 1000 translated sentences is 32,596 in total; the number of tokens in 1000 translated parse-then-generated sentences is 27,958; for the 1000 originally English sentences the total number of tokens is 28,436; after parsing-then-generating the total number of tokens in the 1000 originally English sentences is 25,499.

The \emph{proportion} of each POS tag in the dataset can be seen in \cref{tab:unigram_pos}. For three noteworthy tags which can predict whether a text is translated (adpositions, adverbs, and determiners), we see that using AMR as an interlingua decreases the proportion of these tags in the data, which is desired for ADP and ADV (but not for DET).

\begin{table}[tb]
\centering
\small
\begin{tabular}{r  c c c}
 \toprule
& \textbf{ADP} & \textbf{ADV} & \textbf{DET} \\
\midrule
Translations & 0.1129 & 0.0433 & 0.0982 \\
Originally English & 0.1108 & 0.0389 & 0.0984\\
\midrule
MT & 0.1144 & 0.0413 & 0.1004  \\
Para BART & 0.1009 & 0.0457 & 0.0960 \\
Para T5 & 0.1060 & 0.0333 & 0.0958 \\
AMR P-then-G & 0.1103 & 0.0419 & 0.0963 \\
\bottomrule
\end{tabular}
\caption{Relative frequencies of three part-of-speech tags for the original translations and the generated text after application of each of our translationese reduction techniques. The relative frequencies of originally English text are also provided as a baseline.
}
\label{tab:unigram_pos}
\end{table}

Similarly, the T5-based paraphrase model leads to a decrease in all three tags. The BART-based paraphrase model decreases the proportion of adpositions and determiners, but increases the proportion of adverbs.
However, the MT back-translation output shows an increase in adpositions and determiners, and a decrease in adverbs.

\subsection{Discussion of Translationese Metric Results}

Our translationese metrics reveal that back-translation, via machine translation to French and then to English, does not reduce the amount of translationese in the human-translated texts, but rather \emph{exacerbates} it for all three metrics. The T5-based paraphrase model similarly exacerbates the the amount of translationese except for on one metric, which is the unigram bag-of-POS.
As such, quantitatively, we see an indication that these two methods are not effective techniques for translationese reduction.

On the other hand,  
%Comparing the sentence variants along these translationese metrics, 
we find that all three metrics point to a decrease in translationese with our AMR parse-then-generate approach.
The same is true for the BART-based paraphrase model, which effectively reduces the amount of translationese on our metrics and shows the greatest reduction via type-token ratio and count of cohesive markers. 
Both the AMR parse-then-generate approach and the BART-based paraphrase model produce text more like the originally English text per the part-of-speech relative frequencies.
% \nss{discuss relationship to Originally English scores and what it means}

At this point, our results indicate that the BART-based paraphrase model or the AMR parse-then-generate technique may be a successful way to reduce translationese. 
In the next section, we examine whether adequacy and fluency are maintained or sacrificed using these methods of translationese reduction.

\section{Evaluation of Fluency and Adequacy}
\label{sec:fluency_adequacy_evaluation}

\begin{table}[b]
\centering
\small
\begin{tabular}{ r c c c}
 \toprule
& \textbf{BLEURT} & \textbf{COMET}  & \textbf{BERTscore} \\ [0.5ex] 
 \midrule
MT & 80.31 \textsmaller{(1)} & 87.72 \textsmaller{(1)} & 96.00 \textsmaller{(1)} \\
Para BART & 60.05 \textsmaller{(4)} & 74.31 \textsmaller{(4)} & 94.02 \textsmaller{(3)} \\
Para T5 & 70.67 \textsmaller{(3)} & 81.60 \textsmaller{(3)} & 92.79 \textsmaller{(4)} \\
AMR P-then-G & 75.81 \textsmaller{(2)} & 84.95 \textsmaller{(2)} & 94.89 \textsmaller{(2)} \\
% Controlled Gen & 25.85 \textsmaller{(5)} & 36.54 \textsmaller{(5)} & 79.82 \textsmaller{(5)} \\
 \bottomrule
\end{tabular}
\caption{Average BLEURT, COMET, and BERTscore percentages and rankings (in parentheses) for the 1000 generated sentences from each of our three techniques for translationese reduction, compared against the original sentences as references.}
\label{tab:auto_adequacy}
\end{table}

\begin{table*}[htb]
\centering
\small
\begin{tabular}{| c | c | c |}
 \hline
 Difference & Original (Translationese) Sentence & Sentence after parse-then-generate \\
 \hline
 Conciseness & 
 \makecell{
``Mr President, I would firstly like to congratulate\\ the rapporteur, Mr Koch, on his magnificent work\\ and his positive cooperation with the Commission\\ with regard to improving the texts and presenting\\ this report and this proposal.''} & \makecell{``First, I would like to congratulate Mr Koch\\ for his magnificent work and his positive\\ cooperation with the Commission in \\improving the texts and presenting this\\ report and this proposal.''} \\
 \hline
 Cohesive Markers & \makecell{``Most people, \textbf{however}, would like to live in\\ the area in which they were born and raised,\\ if  they were given the chance to, \\\textbf{in other words}, if there was work there .''} & \makecell{``\textbf{But} if given the chance to do that \\ (work there), most would like to live in \\the area where they were born and raised.''} \\
\hline
Word Order & \makecell{``We note, first of all, that the committee considers\\ the data, as presented in the Commission's annual\\ report, to be in too aggregated a form to enable\\ an in-depth evaluation of state aid policy which is\\ simultaneously legitimate, sensitive to national\\ interests and extensive in terms of compliance\\ with the rules of competition, pursuant to the\\ actual terms of the Treaty.''} & \makecell{``First of all, we note that the committee considers\\ the data presented in the Commission's annual\\ report too aggregated to enable an in-depth\\ evaluation of a legitimate state aid policy that\\ is sensitive to national interests and is extensive\\ in terms of compliance with competition rules\\ within the actual terms of the Treaty.''} \\
\hline
\end{tabular}
\caption{Examples of each of the three main differences we note in sentences before and after applying our AMR parse-then-generate method. The cohesive markers are bolded in the respective row.
% \sw{TO DO: bold the relevant portions} \sw{choose a shorter example for word order}
}
\label{tab:amr_examples}
\end{table*}

\begin{table*}[htb]
\centering
\small
\begin{tabular}{| c | c |}
 \hline
Original (Translationese) Sentence & After BART-based paraphrase model \\
 \hline
 \makecell{
``Although there are now two Finnish channels and one Portuguese one, there is still\\ no Dutch channel, which is what I had requested because Dutch people here like to\\ be able to follow the news too when we are sent to this place of exile every month.''} & \makecell{``Although there are now two Finnish\\ channels and one Portuguese one,\\ there is still no Dutch channel.''} \\
 \hline
 \makecell{``Madam President, the presentation of the Prodi Commission's political programme\\ for the whole legislature was initially a proposal by the Group of the Party of\\ European Socialists which was unanimously approved by the Conference of\\ Presidents in September and which was also explicitly accepted by President 
 Prodi,\\ who reiterated his commitment in his inaugural speech.''} & \makecell{``Madam President, the\\ presentation of the Prodi\\ Commission's political programme\\ for the whole legislature.''} \\
\hline
\end{tabular}
\caption{Examples of brevity enforced by the BART-based paraphrase model, with the first example showing acceptable omission, and the second example demonstrating undue omission (with the sentence being incomplete).
}
\label{tab:bart_examples}
\end{table*}

Having established that macro-level indications of translationese are lessened by using either the BART-based paraphrase model or AMR as an interlingua, we now examine the quality of the generated sentences through the lenses of fluency and adequacy\slash meaning preservation.
We report automatic metrics as well as results of a human evaluation study.

\subsection{Automatic Metrics for Meaning Preservation}
\label{ssec:automatic_adequacy}

We use three metrics to automatically calculate meaning preservation via semantic similarity to the reference: BLEURT \citep{sellam-etal-2020-bleurt}, COMET \citep{rei-etal-2020-comet}, and BERTscore \citep{zhang2019bertscore}.
The BERTscore version that we use relies on \texttt{roberta-large}. For COMET, we use the default \texttt{unbabel-comet} model.

As seen in \cref{tab:auto_adequacy}, across all three metrics, MT back-translation has the highest semantic similarity score.
This technique still fails to reduce translationese in the text (per \cref{sec:metrics}).
%, it is able to preserve meaning (when assessed through these three automatic metrics).

The AMR parse-then-generate scores come in second highest for all three metrics. 
The improved naturalness of the AMR parse-then-generate output is also evident when examining input\slash output pairs.
Three major differences we observed after applying the AMR parse-then-generate techniques include (1)~change in word order, (2)~reduction in cohesive markers, and (3)~added conciseness. 
An example of each of these three differences can be seen in \cref{tab:amr_examples}.

Both paraphrase models show substantially decreased semantic similarity, suggesting they may not accurately convey the meaning of the original sentence.
Even the BART-based paraphrase model, which effectively reduced translationese across all three translationese metrics, suffers from low automatic metric scores, reaching a BLEURT score of 60.05 and a COMET score of 74.31. The BART-based paraphrase model has a BERTscore higher than the T5-based paraphrase model, though all four of the BERTscores are quite high and close to each other. 
%These results raise concern for using the BART-based paraphrase model towards translationese reduction, as adequacy is paramount to the task. 
The low scores are likely due to the the paraphrase models emphasizing brevity so much that key information is being discarded. For example, the first item in \cref{tab:bart_examples} shows an acceptable form of brevity, where the omitted content is not essential to reflecting the meaning of the original sentence, whereas the second example unduly cuts out relevant content and is not a complete sentence.
The average sentence length for the BART-based paraphrase model is 15.07 tokens, whereas the average sentence length for the original (translationese) sentences is 31.33 tokens.\footnote{The average sentence length for the AMR parse-then-generate approach is 24.52 tokens; average sentence length for the T5-based paraphrase model is 22.42; average sentence length for the MT back-translation approach is 27.35.}

Thus, the AMR-based technique strikes the best balance between translationese reduction and meaning preservation when assessed via automatic metrics.

\subsection{Human Evaluation}
\label{ssec:human_evaluation}

% Inspecting the output also uncovers adequacy errors
% due to noise in the AMR parsing and generation steps. In one sentence, polarity was affected (``Although there are now two Finnish channels and one Portuguese one, there is still no Dutch channel, which is what I had requested'' $\rightarrow$ ``I have requested that there still be no Dutch channels, although there are two Finnish channels and one Portuguese one''). In another more egregious error, a percentage was changed (``Furthermore, three quarters of our  farm workers [...]'' $\rightarrow$  ``And 34\% of our farm workers [...]''). 
% Such semantic changes are rare.
% We find that, on the whole, our approach 
% preserves meaning while mitigating the amount of translationese. One way to quantify meaning preservation is with semantic similarity: the BERTscore \citep{zhang2019bertscore} between our output and the original translationese sentences is 0.946. 

%In addition to performing automatic evaluations of semantic similarity, 
Finally, we assess adequacy and fluency of the system output through a human evaluation study.
We collect two judgments per item on 75~sets of items, where each set of items consists of all system outputs associated with one original translationese sentence. For adequacy, there were five sentences per item, and for fluency there were six sentences per item, because the original text was also judged.
% 1,650 total judgments (75 x 5 = 375 adequacy judgments, doubly annotated = 750, plus 75 x 6 = 450 fluency judgments doubly annotated, equals 900)
In total, this amounts to 1,650 total judgments (75 $\times$ 5 = 375 adequacy judgments, doubly annotated = 750, plus 75 $\times$ 6 = 450 fluency judgments doubly annotated, equals 900).

12~annotators participated in total and each annotator judged 25~sets. Adequacy and fluency judgments were collected separately and by different annotators. All annotators were either Computer Science or Linguistics graduate students, and all annotators of fluency were native speakers of English. The order of the system output was randomized, such that no individual system would always appear first in the survey. 

The annotators were asked to judge fluency on a scale from 1--4 and adequacy on a scale from 1--4 in reference to the original translationese-afflicted sentence. For fluency, a score of 1 corresponds with text which is ``nonsensical'', a score 2 is assigned for text which is ``poor'' and has many errors which make the text hard to understand, a score of 3 indicates that the quality of the text is ``good'' and largely understandable with few errors, and a score of 4 is for ``flawless'' text---perfectly formed English with no mistakes.
For adequacy, text which has ``no meaning preservation'' and is completely unrelated to the reference receives a score of 1, text which exhibits ``some meaning preservation'' corresponds with a score of 2, text which has ``most'' of the same meaning as the reference gets a score of 3, and a score which conveys ``all'' of the same meaning receives a scores of 4.
% \nss{put screenshots in the appendix showing the wording of questions} \sw{the question wording was influenced by google best practices so i don't want to accidentally publish something that may be internal-only}

The results of this study can be seen in \cref{tab:human_judgments}.
Inter-annotator agreement via Spearman’s correlation is 0.5 for both fluency and adequacy, suggesting moderate agreement, and the automatic metrics of fluency and adequacy show the same pattern as the human evaluation.

\begin{table}[t]
\centering
\small
\begin{tabular}{ r c c}
 \toprule
& \textbf{Avg Adequacy} & \textbf{Avg Fluency} \\ [0.5ex] 
 \midrule
MT & 3.59 \textsmaller{(1)} & 3.35 \textsmaller{(2)} \\
Para BART & 2.45 \textsmaller{(4)} & 1.91 \textsmaller{(5)} \\
Para T5 & 2.97 \textsmaller{(3)} & 3.39 \textsmaller{(1)} \\
AMR P-then-G & 3.34 \textsmaller{(2)} & 2.76 \textsmaller{(4)} \\
% Controlled Gen. & 1.10 \textsmaller{(6)} & 1.30 \textsmaller{(6)} \\
Originals & N/A & 3.19 \textsmaller{(3)} \\
 \bottomrule
\end{tabular}
\caption{Average adequacy and fluency scores (and their rankings in parentheses) from our human evaluations on 75 sentence sets, comprising 1,650 total judgments.
% (75 x 5 = 375 adequacy judgments, doubly annotated = 750, plus 75 x 6 = 450 fluency judgments doubly annotated, equals 900). 
Originals were used as references in adequacy judgments.}
\label{tab:human_judgments}
\end{table}

Generally, we find that the MT back-translation and AMR parse-then-generate approaches achieve the highest adequacy, as indicated by the automatic metrics (\cref{tab:auto_adequacy}).
While the T5-based paraphrase model output is highly fluent, its adequacy is low, and does not effectively reduce translationese per our prior translationese metrics (\cref{sec:metrics}). The AMR parse-then-generate output suffers from a lower degree of fluency than the MT back-translation and T5-based paraphrase approaches, though the AMR-based output is still sufficiently fluent (as judged qualitatively and via automatic metrics) to ensure readability and meaningfulness. Further progress on AMR-to-text generation models will enable more fluent output.

Additionally, it is worth noting that fluency is low in the human evaluation even for the human-produced originals. As indicated in annotators' comments, this low fluency is likely due to the domain being European Parliament proceedings, which can be complicated for lay people to comprehend (even as our fluency annotators were all native speakers of English).

\subsection{Tradeoff between Translationese Reduction and Maintaining Fluency\slash Adequacy}
\label{ssec:tradeoff}
% \sw{Make point about tradeoff more prominent—move before conclusion?}

Our results reveal the tradeoff between reducing the presence of translationese, while maintaining fluency and adequacy.
Given that the goal is translationese reduction in text, our AMR-based approach is best suited for this task. 
Across three metrics, we demonstrate the utility of AMR in making translated texts more similar to originally English texts.
The AMR parse-then-generate method doesn't perfectly maintain fluency, but based on the automatic metrics and human judgments, still achieves fluency only a bit below that of the original human utterances. 
% fluency at the item level is not the main point of translatioense.
Importantly, adequacy is maintained by the AMR parse-then-generate approach, indicating that information is not lost by using AMR as an interlingua, and suggesting that humans perhaps disprefer the phrasing of the AMR output, while it is still accurately conveying the necessary information.
% \nss{this feels long-winded for a conclusion. are there points here that can be moved to the discussion?} \sw{I don't see anything that can be obviously moved to one of the results sections, because there isn't really a final ``discussion'' section}

\section{Related Tasks}
% \nss{this is loosely related work. I'd move it to a section right before the conclusion (and condense if necessary)}
% \sw{can condense here if necessary}
Our task of translationese reduction on human-translated text is related to the tasks of style transfer, grammatical error correction, paraphrase generation, text simplification, and automatic post-editing, because all of these aim to edit text after generation or produce new text with the same meaning as other text. 

Style transfer and grammatical error correction aim to control features of generated text. Style transfer can control, for example, whether the style is modern or classical, honorific or non-honorific, or conforms to European or Brazilian Portuguese \citep{wang-etal-2023-controlling}.
Style transfer considers what type of style the generated\slash translated text takes on, not whether the text has broader features of translationese.
% \nss{long sentence---break it down}
Recent work on style transfer has leveraged AMR as an interlingua \citep{jangra-etal-2022-star}.
Grammatical error correction removes errors from text \citep{wang2021comprehensive} and aims for fluency, but even error-free fluent text can exhibit features of translationese, such as the source language shining through (\cref{sec:fluency_adequacy_evaluation}).
% \nss{GEC aims for fluency, but even fluent text can bear the hallmarks of translationese. (explain)}

Paraphrase generation is the task of producing sentences which have essentially the same meaning but different syntax and\slash or word choice \citep{zhou-bhat-2021-paraphrase}.
% \nss{why detection? isn't paraphrase generation more relevant?}
\Citet{huang-etal-2022-unsupervised-syntactically} use AMR to control the semantics of generated paraphrases.
% \sw{code is unavailable even upon request.}
Similarly, paraphrase detection determines whether one sentence has the same meaning as another. 
\Citet{issa-etal-2018-abstract}
% An approach using AMR for paraphrase detection 
combine AMR parses with latent semantic analysis to compare two sentences and identify whether they are paraphrases. 

Text simplification aims to make text more readable and easier to process \citep{chandrasekar-etal-1996-motivations}. Research on this task has employed a variety of neural models \citep{nisioi-etal-2017-exploring}.

% \nss{rephrased:}
While in this work we focus on adjusting human translations, a related goal might be to reduce translationese in machine translation output. 
Reducing translationese in machine translations is distinct from automatic post-editing,
\footnote{Human post-editing involves humans looking at generated translations and altering them for increased fluency\slash quality; automatic post-editing aims to automate this process \citep{do2021review}.}
% Automatic post-editing is also distinct from our task, 
not only because modern automatic post-editing requires the use of both the source sentence and the translation (while we do not assume access to any information other than the translation we aim to alter) \citep{chollampatt-etal-2020-automatic}, but more importantly because post-editing exhibits a heightened degree of translationese \citep{toral-2019-post}.
% While post-editing is a worthwhile endeavor to improve MT output, post-editing (both human and automated) demonstrates an even greater degree of the phenomenon we are trying to diminish

Other research at the intersection of AMR and translation has used AMR to improve neural machine translation, unrelated to translationese \citep{song-etal-2019-semantic,nguyen2021improving,li-flanigan-2022-improving}, and framed AMR generation as a machine translation problem \citep{pust-etal-2015-parsing,castro-ferreira-etal-2017-linguistic}.

\section{Conclusion}

% In using AMR as an interlingua to abstract away from surface-level features of the text, we are able to mitigate the effect of translationese in translated texts. 

In this work, we investigated the task of translationese reduction and introduced three methods for this task. 
%Though prior work has studied the characteristics and impact of translationese, approaches to reducing the amount of human translationese is an understudied task---one which we tackle in this work.
Our automatic metrics of translationese indicate that the task of translationese reduction is complicated, because we want translationese to be reduced without sacrificing fluency or adequacy (this tradeoff is discussed in \Cref{ssec:tradeoff}). Overall, we find that using AMR as an interlingua aids in translationese reduction. By contrast, a BART-based paraphrase model is even more effective at reducing translationese, but dramatically over-summarizes, severely harming adequacy and fluency.
The T5-based paraphrase model and MT back-translation approach do not show promise for this task.

Our findings suggest that translationese reduction could be performed as an additional step after translating to make the text more like originally English text, and provides further indication that AMR can serve as an interlingua for a range of tasks which require abstracting away from specific language features \citep[cf.][]{xue-etal-2014-interlingua,wein-etal-2022-effect,song-etal-2019-semantic,li-flanigan-2022-improving}.
% Despite much work on text-to-AMR parsing and AMR-to-text generation, there is of course some amount of error introduced in our method by parsing and generating; future work on AMR-to-text generation would likely lead to increased fluency.
% We expect that our results will be further improved through continued work on AMR-to-text generation and text-to-AMR parsing. Our current experimentation is also exclusive to English. 
% Given that AMR (and parsers/generators for it) has also been adapted to a number of languages other than English, making it possible to apply the same technique to different types of texts affected by translationese \citep{wein-schneider-2022-accounting}, it would be interesting in future work to adapt this methodology to other languages.

\section*{Limitations}
Despite much work on text-to-AMR parsing and AMR-to-text generation, there is of course some amount of error introduced in our parse-then-generate method. We find in our results that the meaning is preserved, and while fluency is a bit lower, additional progress on AMR-to-text generation research will likely enable further fluency in the end result of using AMR as an interlingua.

Future work may explore the applicability of our methods to languages other than English and additional domains.
% Our current experimentation is exclusive to English;  
% %Though the metrics and AMR parsing and generation techniques are in principle applicable to other languages, 
% the effectiveness of AMR at mitigating translationese in other languages is yet to be seen.
Further, because we have used European Parliament data in this experimentation, all of the source languages are European languages,
% Future work should explore the effectiveness and quality of this approach when using English translated from different source languages, in particular 
and translationese has different features depending on the source language \citep{koppel-ordan-2011-translationese}.
AMR (and parsers\slash generators for it) has also been adapted to a number of languages other than English, so in principle it is possible to apply the same technique to different types of texts affected by translationese \citep{wein-schneider-2022-accounting}.
%it would be interesting in future work to adapt this methodology to other languages.
%
While we have not yet examined the downstream effect of applying our approach, this would be a promising avenue for future work.

% Incorporation into downstream applications
% would be helpful in determining the utility and scope of our approach.  Future work might include producing embeddings of the generated sentences and seeing if that improves performance on downstream tasks, as was explored in \citet{dutta-chowdhury-etal-2022-towards}.

% \section*{Ethics Statement}

\section*{Acknowledgements}
This work is supported in part by a Clare Boothe Luce Scholarship and NSF award IIS-2144881. We thank Sireesh Gururaja and anonymous reviewers for their feedback.
Thank you to the following people for supplying human judgments of fluency and adequacy: Chao-Chin Liu, Sajad Sotudeh, Jianan Su, Ke Lin, Xiulin Yang, Laasya Bangalore, Devika Tiwari, Autumn Toney-Wails, Rahel Fainchtein, Ryan Wails, Samuel King, and Thomas Lupicki.

% Entries for the entire Anthology, followed by custom entries
\bibliography{anthology,custom}
\bibliographystyle{acl_natbib}

% \appendix

\end{document}